\documentclass{article} 
\usepackage{iclr2025_conference,times}

\usepackage{amsmath,amsfonts,bm}

\usepackage{hyperref}
\usepackage{url}
\usepackage[T1]{fontenc}
\usepackage{comment}
\usepackage{latexsym}
\usepackage{amssymb}
\usepackage{amsmath}
\makeatletter
\@ifundefined{proof}{\usepackage{amsthm}}{}
\makeatother
\usepackage{booktabs}
\usepackage{enumitem}
\usepackage{graphicx}
\usepackage{algorithm}
\usepackage{algpseudocode}
\usepackage{array}
\usepackage{multirow}
\usepackage{footmisc} 
\usepackage{hyperref} 
\usepackage{lipsum} 
\usepackage{graphicx}

\def\eqref#1{equation~\ref{#1}}

\def\1{\bm{1}}

\DeclareMathAlphabet{\mathsfit}{\encodingdefault}{\sfdefault}{m}{sl}
\SetMathAlphabet{\mathsfit}{bold}{\encodingdefault}{\sfdefault}{bx}{n}

\title{Enhancing Interpretability in Generative  AI Through Search-Based Data Influence\\ Analysis}

\author{
Theodoros Aivalis\textsuperscript{1,2}\thanks{ORCID: 0009-0005-4452-9402}, 
Iraklis A. Klampanos\textsuperscript{1}\thanks{ORCID: 0000-0003-0478-4300}, 
Antonis Troumpoukis\textsuperscript{2}\thanks{ORCID: 0000-0003-1078-8121}, 
Joemon M. Jose\textsuperscript{1}\thanks{ORCID: 0000-0001-9228-1759} \\
\textsuperscript{1}University of Glasgow, UK \\
\textsuperscript{2}National Centre for Scientific Research ``Demokritos'', Greece \\
\thanks{Corresponding Author:  Theodoros Aivalis (\texttt{teoaivalis@iit.demokritos.gr})}
}

\iclrfinalcopy 
\begin{document}

\maketitle

\begin{abstract}
\label{subsec:abstract}
Generative AI models offer powerful capabilities but often lack transparency, making it difficult to interpret their output. This is critical in cases involving artistic or copyrighted content. This work introduces a search-inspired approach to improve the interpretability of these models by analysing the influence of training data on their outputs. Our method provides observational interpretability by focusing on a model's output rather than on its internal state. We consider both raw data and latent-space embeddings when searching for the influence of data items in generated content. We evaluate our method by retraining models locally and by demonstrating the method's ability to uncover influential subsets in the training data. This work lays the groundwork for future extensions, including user-based evaluations with domain experts, which is expected to improve observational interpretability further.

\end{abstract}

\section{Introduction}
\label{subsec:introduction}

The rapid growth of deep learning across diverse fields, from art to science, has raised significant ethical, legal, and policy concerns. A key issue is the use of copyrighted content to train generative models, with lawsuits claiming that such training, when done without permission, violates copyright laws. Possible outcomes of these lawsuits include significant financial penalties and sometimes orders to delete these models~\citep{ai-copyright-cases}. In response to these challenges, regulatory bodies and international organisations have proposed various frameworks to mitigate potential risks. The European Union has introduced the AI Act~\citep{euaiact}, while the UK has outlined its National AI Strategy~\citep{uk_aistrategy}. Additionally, UNESCO~\citep{unesco_ai} has established principles for responsible AI development, and the European Group for AI Regulation\footnote{\url{https://www.egair.eu/}, as viewed February 2025.} advocates for AI technologies that respect privacy and copyright.

Despite these regulatory efforts, the lack of transparency in generative models remains a major challenge. Most large-scale generative models operate as black boxes, making it difficult to trace how training data influences their outputs. Even open-source models, while more accessible, are typically too large to fully reproduce results, further complicating interpretability. Existing methods for data influence analysis, such as influence functions and training data attribution, primarily focus on parameter-based explanations and require access to the model’s internal gradients, making them impractical for black-box generative models.

In this paper, we propose a  search-based data influence analysis  framework that enhances the interpretability of  generative models  by directly  linking training data to generated outputs . Our method consists of two main steps:
\begin{enumerate}
    \item Retrieving relevant training samples based on the user’s input prompt.
    \item Performing a comparative analysis between these samples and the generated output using raw data similarity and latent-space embeddings.
\end{enumerate}

By leveraging the user’s prompt into the retrieval phase, our approach ensures that the selected training samples align textually with the input, capturing essential characteristics at the word level. The second phase then systematically compares these retrieved samples with the generated output using both raw data and embedding-based similarity measures. This structured two-step process enables a principled means of tracing influential training data, even in black-box generative models, thereby enhancing interpretability.

The main contributions of this paper are as follows:
\begin{itemize}
    \item We introduce a novel, model-agnostic search-based data influence analysis framework that provides observational interpretability for generative models by linking generated outputs to training data. Our hybrid retrieval-comparison approach ensures both efficiency and robustness by combining text-based and image-based similarity.
    \item We demonstrate the effectiveness of our framework through experiments on both locally trained and large-scale generative models, showing that removing influential training samples results in a measurable reduction in similarity between generated and training data.
    \item Our method has practical implications for copyright tracking and dataset transparency, providing a possible tool for identifying  which training data influenced a generated output.

\end{itemize}

The rest of this paper is structured as follows: Section 2 reviews related work on interpretability and data influence analysis. Section 3 presents our search-based framework and its theoretical foundation. Section 4 reports experimental results on both locally trained and large-scale generative models. Finally, Section 5 concludes with key findings and future directions.

\section{Related Work}
\label{subsec:related work}

As Generative AI evolves, the need for improved interpretability increases.
Visualisation techniques like t-SNE~\citep{t-SNE} allow for the projection of complex, high-dimensional datasets into lower dimensions for easier interpretation. Local methods such as LIME~\citep{lime} offer feature-level insights, contributing to the interpretability of models. Visualising transformer attention~\citep{attention} and Latent Space Analysis provide further clarity into the decision-making processes of complex models.

A key aspect in interpretability is understanding how training data influences model behaviour.
Traditional Training Data Attribution (TDA) approaches estimate influence by modifying the dataset and analysing the impact on model performance. For instance, Leave-One-Out Influence (LOO)~\citep{black2021leaveoneoutunfairness} quantifies the effect of individual data points by retraining the model without them. Downsampling~\citep{downsampling} evaluates the influence of data subsets, balancing efficiency and accuracy. Influence functions~\citep{koh} trace predictions to key training samples, helping identify instances that shape model behaviour. These methods provide insights into model sensitivity but often require gradient access and significant computational resources. Shapley Values~\citep{pmlrshapley} offer a game-theoretic perspective on TDA by assigning contribution scores to training samples, though they are computationally expensive.
Bayesian approaches to TDA have been introduced to address the limitations of deterministic attribution methods. Nguyen et al.~\citep{bayesianapproach} propose a probabilistic framework, where influence scores are treated as random variables. This approach accounts for uncertainty in attribution estimates, recognising that model initialisation and data noise can significantly affect influence scores.

A significant challenge about interpretability of generative models is the lack of ground truth for explanations. Unlike supervised learning, where ground truth allows for objective evaluations, explanations are inherently subjective and context-dependent. Without a universally agreed-upon standard, numerical comparisons across datasets may fail to capture the strengths and limitations of different methods. We highlight this by presenting a qualitative comparison of existing baseline methods in data influence analysis alongside our proposed approach. This comparison highlights the strengths, weaknesses, and applicability of each method in diverse scenarios.

\begin{table}[ht]
\centering
\caption{Comparison of methods for data influence analysis and interpretability.}
\label{tab:method_comparison}
\resizebox{\textwidth}{!}{%
\begin{tabular}{c c c c c c}
\toprule
\textbf{Method}                  & \textbf{Focus}          & \textbf{Model}         & \textbf{Computational}  & \textbf{Task}     & \textbf{Black-Box} \\ 
                                 &                         & \textbf{Type}          & \textbf{Cost}           & \textbf{Type}     & \textbf{Compatible} \\ 
\midrule
Influence Functions              & Model Parameters    & Any  & High                    & Classification,   & No                  \\ 
                                 &                         &                        &                         & Regression        &                     \\ 
Shapley Values                   & Training Dataset & Any                    & Very High               & Classification,   & Yes                 \\ 
                                 &                         &                        &                         & Regression,       &                     \\ 
                                 &                         &                        &                         & Generation        &                     \\ 
\midrule
Leave-One-Out                    & Training Dataset        & Any                    & Very High               & Classification,   & Yes                 \\ 
                                 &                         &                        & (Retraining)            & Regression,       &                     \\ 
                                 &                         &                        &                         & Generation        &                     \\ 
Baysian TDA                      & Training Dataset & Any                    & High                    & Classification,   & No                  \\ 
                                 &                         &                        &                         & Regression        &                     \\ 
\midrule
Latent Space Analysis            & Latent Representations  & Embedding Models       & Medium                  & Classification,   & No                  \\ 
                                 &                         &                        &                         & Generation        &                     \\ 
Attention Visualisation          & Attention Weights       & Transformers,          & Medium-High             & Text/Image        & No                  \\ 
                                 &                         & Attention Models       &                         & Tasks (Any type)  &                     \\ 
\midrule
\textbf{Our Method}              & \textbf{Training Dataset} & \textbf{Any}          & \textbf{Low}            & \textbf{Any} & \textbf{Yes}       \\ 
\bottomrule
\end{tabular}%
}
\end{table}

Table \ref{tab:method_comparison} summarises this comparison, detailing how traditional methods like Influence Functions and Leave-One-Out focus on individual data points but struggle with computational scalability in large models. Shapley Values, although black-box compatible, are computationally prohibitive due to their factorial complexity. Latent space analysis methods and attention visualisation focus on specific aspects of the model, such as representations or attention patterns, but do not directly address data influence on predictions.
Our method balances computational efficiency with robustness, offering insights into data influence without requiring costly retraining or complex approximations. Its compatibility with black-box models and focus on generation tasks make it applicable to various scenarios. While existing methods have limitations, our approach provides a scalable and interpretable way to explore the relationship between training data and generated outputs.

Recent advances in retrieval-augmented techniques aim to improve model interpretability and efficiency. Methods focusing on enhancing model reasoning and retrieval-based improvements, like Chain of Thoughts(CoT)~\citep{chainofthoughts} enhance interpretability by breaking down complex reasoning into intermediate steps. Approaches as above, such as Mixture of Experts(MoE)~\citep{mixturehugging} and Retrieval-Augmented Generation(RAG)~\citep{surveyrag} improve efficiency and accuracy by dynamically selecting parameters or retrieving external knowledge to enhance model performance and transparency.

While all these methods advance interpretability, they tend to overlook the ethical and legal challenges that large generative models pose to artists and creators. These challenges stem from often disregarding the use of copyrighted material during training or fine-tuning. Furthermore, the sheer size and deliberate opacity of the methodology, training dataset, and model make it difficult for artists to interrogate and challenge their use. In this context, methods for tracing the influence of training samples remain unexplored as potential tools for ensuring ethical AI practice and attribution.
Our method addresses this gap by directly linking training samples to generated outputs through a two-step search-based framework. Unlike methods relying on internal parameters or those that focus only on input and the generated output, we correlate outputs with training data through features that can be captured from textual and visual representations, making our approach suitable for black-box models. Combining textual and visual analyses, our method identifies semantically and visually similar training samples, addressing copyright concerns. This leads to a richer and more comprehensive understanding of the data's influence on generated outputs.

\section{Proposed Method}
\label{sec:proposed_method}

In this paper, we propose a data-centric approach to enhance the interpretability of generative models, addressing challenges such as potential copyright violations. Existing methods, such as Influence Functions and Training Data Attribution, primarily assess parameter shifts without fully explaining the generation process. In contrast, our method links training samples to generated outputs by analysing the content and structure of the data, thereby improving observational interpretability.

To formalise our method, we model the generative process as follows:

\[
y = g(x) + \epsilon,
\]

where \( y \) is the generated output for a given user-provided prompt \( x \), \( g(x) \) represents the model’s learned function, and \( \epsilon \) accounts for variability and imperfections in generation.

To quantify the contribution of (m) individual training samples to the generated output, we define the data influence function:

\begin{equation}
\text{Data Influence}(x, y) = \sum_{i=1}^{m} \alpha(y, y_i) \cdot B(x, x_i),
\end{equation}

where:
\begin{itemize}
    \item \( B(x, x_i) \) is a binary filter function ensuring that only textually relevant training samples contribute.
    \item \( \alpha(y, y_i) \) is the kernel-based influence weight of each training sample, inspired by non-parametric regression techniques such as the Nadaraya-Watson estimator~\citep{nadaraya}.
\end{itemize}

To filter training samples based on textual similarity to the prompt \( x \), we define:

\begin{equation}
B(x, x_i) = 
\begin{cases} 
1, & \text{if } x_i \text{ is among the} \text{ most textually similar samples}, \\ 
0, & \text{otherwise}.
\end{cases}
\end{equation}

The kernel-based weight \( \alpha(y, y_i) \) is computed as:

\begin{equation}
\alpha(y, y_i) = \frac{K(y, y_i)}{\sum_{j=1}^{m} K(y, y_j)},
\end{equation}

ensuring that the total influence is normalised:

\begin{equation}
\sum_{i=1}^{m} \alpha(y, y_i) = 1.
\end{equation}

The visual similarity kernel \( K(y, y_i) \) is defined using cosine similarity:

\begin{equation}
K(y, y_i) = \frac{\text{emb}(y) \cdot \text{emb}(y_i)}{\|\text{emb}(y)\| \cdot \|\text{emb}(y_i)\|}.
\end{equation}

where:
\begin{itemize}
    \item \( \text{emb}(y) \) and \( \text{emb}(y_i) \) are embedding representations of the generated and training images.
    \item The kernel assigns higher weights to visually similar training images.
\end{itemize}

\begin{figure}[htbp]
    \centering
    \includegraphics[width=1\textwidth]{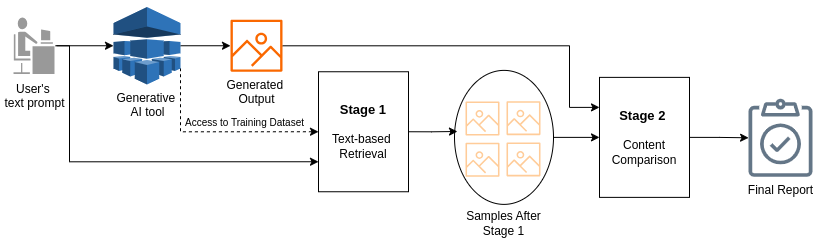}  
    \caption{Overview of the proposed pipeline. \textbf{Phase one} performs text-based retrieval by comparing the \textbf{user's prompt} with the text descriptions of the training samples. \textbf{Phase two} refines the results by comparing the retrieved samples with the \textbf{generated output} with both textual and visual features.}
    \label{fig:proposed_method}
\end{figure}

We approximate our theoretical definition through a two-step process, as outlined in Figure~\ref{fig:proposed_method}.
First, the retrieval process uses the text descriptions of training samples and the user's prompt to efficiently identify relevant samples. These text descriptions capture important characteristics, both for training the generative model and for guiding the retrieval process. 
For the text-based retrieval phase of our method, we choose TF-IDF (Term Frequency-Inverse Document Frequency) \footnote{\url{https://scikit-learn.org/stable/modules/generated/sklearn.feature_extraction.text.TfidfVectorizer.html}, as viewed February 2025} over embeddings due to its interpretability and suitability for keyword-based retrieval tasks. 
TF-IDF quantifies the importance of a term in a document by considering both its frequency within the document and its rarity across multiple documents. In our framework, document frequency was computed across training samples.

After retrieving these candidate samples, a more detailed comparison is conducted to evaluate the similarity between the retrieved samples and the generated output. By comparing the features of the generated output with those of the retrieved samples, the goal is to refine the selection and clarify how specific training data influence the generated content. This two-phase approach ensures that the retrieved samples are relevant in terms of both textual content and visual alignment with the generated output, enhancing model interpretability.

Text-based retrieval can also play an important role in identifying potential copyright issues. When users provide prompts, they may intentionally or unintentionally guide the generation process to create outputs influenced by copyrighted material. By retrieving and analysing the training samples most relevant to the prompt, our method serves as a useful first step for addressing copyright concerns due to the use of generative AI systems. 
However, initially relying on text-based retrieval potentially has limitations, namely that it assumes high recall when retrieving potentially influential items. This is an assumption the investigation of which is left as future work. At this stage, we take a meaningfully large number of retrieved items to ensure we have enough items to work with without jeopardising efficiency i.e., without having to generate embeddings for the entire dataset. The cut-off of the text-based search results is selected experimentally, based on dataset characteristics. 

The second phase of our method improves on the initial retrieval results. By employing image-based similarity, this  step acts as a further filtering mechanism, removing less relevant samples and improving overall precision ~\citep{introductiontoIR}.

\section{Experiments and Method Analysis}
\label{subsec:experiments}

In this section, we evaluate our method using two setups: (1) a locally trained generative model for controlled analysis and (2) large-scale generative models without access to their training data. We first evaluate retrieval effectiveness in a structured environment, then extend the evaluation by approximating training data influence using publicly available images. The following subsections detail our methodology and evaluation criteria.

\subsection{Experiments on a locally trained model}
\label{subsec:experiments on a locally trained model}

\begin{figure}
    \centering
    \includegraphics[width=1\textwidth]{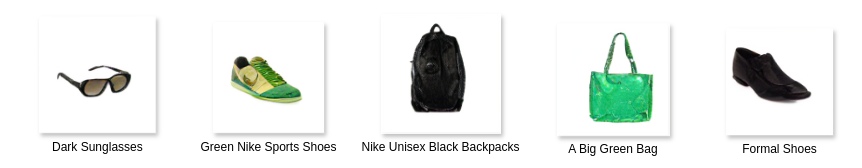}
    \caption{Sample output from the locally trained DALL·E model, to capture key characteristics for detailed analysis.}
    \label{fig:sample_outputs}
\end{figure}

Despite widespread discussions on the importance of open-source LLMs \footnote{E.g. ``Meta’s AI Chief Yann LeCun on AGI, Open-Source, and AI Risk'', \url{https://time.com/6694432/yann-lecun-meta-ai-interview}, as viewed February 2025.}, most generative models remain closed with inaccessible training data. Open-source alternatives, while available, are often too large to reproduce results efficiently, as they require immense computational resources and complex setups. To overcome this problem, we experimented using a locally trained model, which allowed for better control and alignment with our requirements. Specifically, we utilised the Dalle-pytorch package \footnote{\url{https://github.com/lucidrains/DALLE-pytorch}, as viewed February 2025.}
, which replicates OpenAI’s DALL-E model, to perform image generation tasks based on text prompts.
Our methodology comprises two steps: (1) the retrieval of relevant samples based on textual similarity and (2) the comparison of these relevant samples with the generated image to deduce how and which training data has influenced the generated output.
For our experiments, we used an open dataset of approximately 44,000 images depicting fashion items, each accompanied by textual descriptions~\citep{fashionitems}. We use these textual descriptions for efficient training of the local generative model and retrieving relevant samples during the search process. Due to its thematic coherence, this dataset is effective for consistent training as well as it can serve for establishing a solid baseline for experimentation.
Figure~\ref{fig:sample_outputs} demonstrates the ability of the locally trained model to effectively visualise key characteristics of fashion items.
The source code for training the model, along with a variety of generated examples showcasing the capabilities of the created model, can be found in the associated GitHub repository.
\footnote{\url{https://github.com/teoaivalis/Search-Based_Data_Influence_Analysis.git}} 
%

\subsubsection{Text-based Image Retrieval Step}
\label{subsec:Text-based Image Retrieval Step}
During this initial phase we focus on the textual descriptions of the training dataset, which capture important characteristics of the corresponding images. 
This ensures the alignment of the retrieval process with the user's prompt, enabling efficient identification of images that closely match the given text input.

We estimate the relevance between the user's prompt and the training data descriptions by measuring the cosine similarity between TF-IDF representations. Cosine similarity is a widely used metric which has also been found to be effective at similarity-based explanation of machine learning models~\citep{hanawa2021cosine}. This approach is more efficient, as it avoids generating embeddings for the entire dataset, while it is also more readily interpretable.
To refine the retrieval process and focus on a manageable subset, we retained approximately 0.2\% of the full dataset of 44,000 images. This percentage was chosen based on empirical experimentation, where we tested different small subset sizes and evaluated the relevance of the retrieved samples. Through these trials, we observed that retaining 0.2\% of the dataset provided a good balance—ensuring that the subset remained interpretable while still capturing meaningful connections between the retrieved samples and the generated output.

\subsubsection{Image Similarity Analysis Step}
\label{subsec:Image Similarity Analysis Step}
We compared generated images to a subset of the training data, using both raw pixel values and ResNet-50 embeddings to capture low-level and high level image features, respectively. The 10\% most similar training images, based on this combined similarity measure, were defined as the most influential. This ranking provides insights into each training sample's contribution to the generated output.
We selected cosine similarity for both raw images and embeddings, as it effectively handles numerical data, making it well-suited for our approach. This combined approach, using both raw pixel values and embeddings, capture both low-level and high-level image features for a more comprehensive comparison.
In the next steps, we plan to conduct an ablation study to refine the contribution of raw image and embedding similarities, aiming to determine their optimal balance for more precise and informative comparisons.

\subsubsection{Evaluation using the Locally Trained Model}
\label{subsec:method's evaluation}
We evaluated our method's effectiveness by assessing how influential training samples affect image generation, inspired by group influence analysis~\citep{influencesurvey}, which evaluates the collective impact of subsets of training samples on model predictions. 
For each experiment, we generated 15 prompts by randomly sampling and modifying fashion item descriptions from the training data (e.g., altering color, shape, or brand). Influential training samples, identified via text retrieval and image similarity analysis (Sections~\ref{subsec:Text-based Image Retrieval Step} and~\ref{subsec:Image Similarity Analysis Step}), were then "unlearned" by retraining the model without them, using identical hyperparameters. We compared images generated before and after unlearning to a reference image from the official model, using cosine similarity of image embeddings as our primary metric. Consistent SSIM values confirmed that image structure and perceptual quality were preserved, indicating that observed changes resulted from removing influential samples.
Table~\ref{tab:unlearning} shows cosine similarity between reference and generated images, before and after unlearning, including standard deviation and range. The consistent decrease in similarity after unlearning suggests the removed images significantly influenced generation. Post-unlearning, generated images shared fewer features with the reference, indicating influence from other training samples. These results support the potential our method has for tracing the contribution of copyrighted material to model outputs.

\begin{table}[ht]
\centering
\setlength{\tabcolsep}{3pt} 
\renewcommand{\arraystretch}{1.1} 
\begin{tabular}{c cccc c cccc}
\toprule
\textbf{} & \multicolumn{4}{c}{\textbf{Unlearning Experiments 1-5}} & & \multicolumn{4}{c}{\textbf{Unlearning Experiments 6-10}} \\ 
\cmidrule(lr){2-5} \cmidrule(lr){7-10}
          & \textbf{Stage} & \textbf{Mean} & \textbf{Std} & \textbf{Range} & & \textbf{Stage} & \textbf{Mean} & \textbf{Std} & \textbf{Range} \\ 
\midrule
\textbf{} & Before & \textbf{0.546} & 0.051 & 0.465 - 0.606 & & Before & \textbf{0.531} & 0.056 & 0.4525 - 0.614 \\
             & After  & \textbf{0.522} & 0.049 & 0.453 - 0.586 & & After  & \textbf{0.492} & 0.057 & 0.411 - 0.561 \\ 
\cmidrule(lr){1-5} \cmidrule(lr){7-10}
\textbf{} & Before & \textbf{0.537} & 0.043 & 0.477 - 0.596 & & Before & \textbf{0.553} & 0.062 & 0.459 - 0.631 \\
             & After  & \textbf{0.488} & 0.051 & 0.411 - 0.551 & & After  & \textbf{0.518} & 0.077 & 0.41 - 0.616 \\ 
\cmidrule(lr){1-5} \cmidrule(lr){7-10}
\textbf{} & Before & \textbf{0.512} & 0.072 & 0.421 - 0.613 & & Before & \textbf{0.523} & 0.070 & 0.419 - 0.613 \\
             & After  & \textbf{0.481} & 0.057 & 0.411 - 0.569 & & After  & \textbf{0.486} & 0.067 & 0.39 - 0.584 \\ 
\cmidrule(lr){1-5} \cmidrule(lr){7-10}
\textbf{} & Before & \textbf{0.519} & 0.047 & 0.455 - 0.585 & & Before & \textbf{0.544} & 0.068 & 0.451 - 0.633 \\
             & After  & \textbf{0.475} & 0.053 & 0.395 - 0.551 & & After  & \textbf{0.510} & 0.071 & 0.412 - 0.599 \\ 
\cmidrule(lr){1-5} \cmidrule(lr){7-10}
\textbf{} & Before & \textbf{0.550} & 0.056 & 0.466 - 0.622 & & Before & \textbf{0.526} & 0.060 & 0.439 - 0.605 \\
             & After  & \textbf{0.508} & 0.069 & 0.431 - 0.613 & & After  & \textbf{0.504} & 0.058 & 0.424 - 0.585 \\ 
\bottomrule
\end{tabular}

\caption{Analysis of the \textbf{cosine similarity} between generated images before and after unlearning. The table reports \textbf{Mean, Std deviation, and Range of the scores}. A decrease in similarity suggests that the images removed were influential on the image generation.}
\label{tab:unlearning}
\end{table}

\subsection{Experiments with Large-Scale Models}
\label{subsec:experiments with large-scale models}

\begin{figure}
    \centering
    \includegraphics[width=1\textwidth]{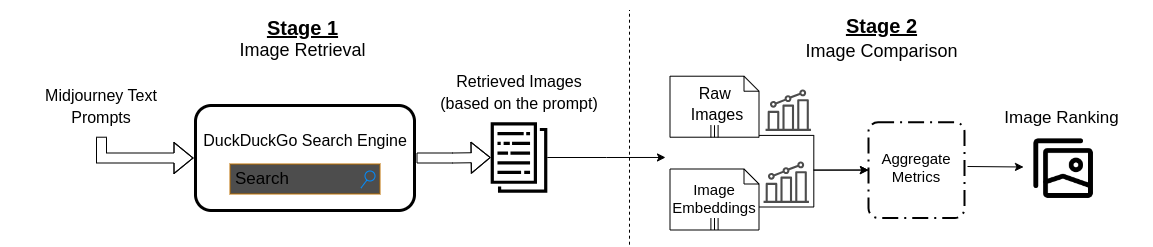} 
    \caption{Comparison of Stable Diffusion generated images with DDG-retrieved images, demonstrating the effectiveness of our retrieval-based method.}
    \label{fig:ddg_experiments}
\end{figure}

Most large-scale generative models remain closed, not providing access to training dataset or training methodology. This lack of transparency highlights the need for experimentation frameworks that can investigate the influence of training data on their outputs without having direct access to the datasets. Here we provide a practical approach for approximating the influence of images accessible on the public Web for image generation.
This is based on the assumption that a substantial portion of training data for these models originates from the public Web, as supported by Larousserie \footnote{\url{https://www.lemonde.fr/en/science/article/2024/07/09/inside-the-secrets-of-generative-ai_6678442_10.html}, as viewed February 2025}.
This methodology serves as an initial step toward adapting our approach for large-scale generative models, which are increasingly integrated into real-world applications.
The process and results are detailed in the following sections and illustrated in Figure~\ref{fig:ddg_experiments}.

\subsubsection{Retrieve Images from the Search Engine}
\label{subsec:retrieve images from the search engine}
We used the Midjourney dataset~\citep{midjourney_data} as queries to a web-based search. This dataset contains approximately 150,000 text-image pairs generated using Stable Diffusion, collected from Discord channels \footnote{\url{https://discord.com/channels/662267976984297473/999550150705954856}, as viewed February 2025.}. To maintain an unbiased search, we excluded prompts with reference images, resulting in 57,000 prompts under 170 characters. For the other 16,000 longer prompts, we simplified them by retaining only nouns and verbs.
The search engine DuckDuckGo \footnote{\url{https://duckduckgo.com}, as viewed February 2025} was selected due to its large index of images. We used DuckDuckGo, in this case that we don't have access to the training dataset, to retrieve 30 images per prompt, as this number provided a good balance between retrieval diversity and computational efficiency. This corresponds to the text-based image retrieval step described in Section ~\ref{subsec:Text-based Image Retrieval Step}.

\subsubsection{Comparison of Retrieved Images}
\label{subsec:compare retrieved images}
We initially compared both raw images and embeddings obtained from a ResNet50 network. Using the generated image as a reference, we compared it to the retrieved ones. The collected metrics were combined following the procedure outlined in Section~\ref{subsec:Image Similarity Analysis Step}.
The results indicate a mean cosine value of \(0.5753\), with a Std deviation of \(0.0546\) and a cosine range of \(0.3536\) to \(0.7934\). These metrics highlight that this experimentation is promising, as the retrieved images are similar to the generated. Examples from the retrieval process, along with the ranking results for selected prompts, are available in our GitHub repository for further exploration.

\section{Conclusions and Future Work}
\label{sec:conclusion_and_future_work}

In this paper, we introduce an approach to enhance the interpretability of generative models based on observations. By combining text-based retrieval with image similarity analysis, we identify training samples that appear to significantly influence the generation process. Through retraining after removing influential samples, we validate our method by observing and quantifying changes in regenerated outputs. 
Generative models often rely on vast datasets sourced from the public Web, raising concerns about transparency and ethical considerations. Our approach highlights the necessity of having access to training datasets or metadata to enable effective influence analysis. Encouraging AI companies to share metadata where possible could foster greater transparency and ethical evaluation in generative AI systems.

Finally, our method is designed to be model-agnostic. Its reliance on textual descriptions and semantic comparisons ensures flexibility, even when metadata characteristics vary significantly. By emphasising generalisability, our approach demonstrates its potential to provide interpretability and influence analysis for a wide range of generative models, regardless of architecture or training dataset. 
While similarity does not always equate to influence, our framework ensures that influential training samples are identified effectively. To further validate this assumption, we will conduct experiments comparing the effects of removing highly similar samples versus those with lower similarity but potential influence. This will refine our method and enhance its robustness in capturing truly influential training data. Overall, our approach contributes to improving transparency in generative AI by providing a structured way to link training data with generated outputs.

In the near future, we intend to expand our approach to enhance its applicability and interpretability across diverse domains and modalities. One direction is to explore additional datasets, such as COCO~\citep{coco_dataset}, Flickr30k~\citep{flickr}, and domain-specific datasets like paintings, to assess the performance and adaptability of our method in different contexts. Experimenting with more specialised datasets will help evaluate the method’s generalisability.
Improving the retrieval process is another priority. We aim to implement scalable retrieval techniques to handle larger datasets more efficiently. Additionally, we will explore automated metadata generation for opaque models, leveraging LLMs or image captioning tools to create structured textual descriptions. This will ensure that our method remains effective even when access to training datasets is limited or unavailable. 

Another promising direction is the integration of user-driven feedback and domain expertise to capture qualitative aspects such as style and context. This will enhance the retrieval relevance and interpretability of the results, particularly in art-related applications. Furthermore, we aim to explore alternative more explainable representations, such as graph-based structures, to replace or complement embeddings. These explainable structures could provide a more intuitive understanding of relationships between training data and generated outputs, particularly to non-technical users.
Another area that could prove useful is to better understand the needs of real-world users. To that end, we plan to conduct surveys with copyright holders, such as photographers, painters, and artists. These insights will help align our method with the requirements of those directly impacted by generative AI, providing practical value.

\subsubsection*{Acknowledgment}
This work has received funding from the European Union’s Digital Europe Programme (DIGITAL) under grant agreement No 101146490.

\bibliography{new_iclr_paper}
\bibliographystyle{iclr2025_conference}

\end{document}